\documentclass[letterpaper, 10pt, conference]{ieeeconf}

\usepackage[left=54pt,top=50pt,right=54pt,bottom=57pt]{geometry}

\usepackage{times}
\usepackage{tabularx}
\usepackage{subcaption}
\usepackage{multirow,multicol, array}
\usepackage[font={small}]{caption}   
\usepackage{graphicx,dblfloatfix}
\usepackage{wrapfig}
\usepackage{siunitx}
\usepackage{soul}

\let\labelindent\relax
\usepackage{enumitem}

\usepackage{amsmath,amsthm,amssymb,amsfonts}
\usepackage[titlenumbered,ruled]{algorithm2e}
\usepackage{textcomp}
\usepackage{acronym}
\usepackage{balance}
\usepackage{mdwmath}
\usepackage{bm}
\usepackage{mdwtab}
\usepackage{array}
\usepackage[usenames,dvipsnames]{color}
\usepackage{eqparbox}
\usepackage{cite}
\usepackage{verbatim}
\usepackage{amssymb}

\usepackage{color}
\usepackage{xcolor}
\usepackage{psfrag}
\usepackage{epsfig}
\usepackage{url}

\usepackage{epstopdf}
\usepackage{booktabs}
\usepackage{blindtext}

\usepackage[colorinlistoftodos,prependcaption,textsize=tiny]{todonotes}

\renewcommand{\emph}{\textit}

\newtheorem*{lemma*}{Lemma}

\newtheorem*{problem*}{Problem}

\makeatletter
\newcommand\fs@spaceruled{\def\@fs@cfont{\bfseries}\let\@fs@capt\floatc@ruled
    \def\@fs@pre{\vspace{5\baselineskip}\hrule height.8pt depth0pt \kern2pt}%
    \def\@fs@post{\kern2pt\hrule\relax}%
    \def\@fs@mid{\kern2pt\hrule\kern2pt}%
    \let\@fs@iftopcapt\iftrue}
\def\maketag@@@#1{\hbox{\m@th\normalfont\normalsize#1}}
\makeatother

\IEEEoverridecommandlockouts
\graphicspath{{images/}}

\begin{document}

	
\title{
Modeling and Trajectory Optimization for Standing Long Jumping \\of a Quadruped with A Preloaded Elastic Prismatic Spine}
	
\author{Keran Ye and Konstantinos Karydis
	\thanks{The authors are with the Dept. of Electrical and Computer Engineering, University of California, Riverside. 
	Email: {\{kye007, karydis\}@ucr.edu}.
	}
	\thanks{
	We gratefully acknowledge the support of NSF under grant \#CMMI-2046270. Any opinions, findings, and conclusions or recommendations expressed in this material are those of the authors and do not necessarily reflect the views of the National Science Foundation.}
}
	
\maketitle
\thispagestyle{empty}


\begin{abstract}
This paper presents a novel methodology to model and optimize trajectories of a quadrupedal robot with spinal compliance to improve standing jump performance compared to quadrupeds with a rigid spine. We introduce an elastic model for a prismatic robotic spine that is actively preloaded and mechanically lock-enabled at initial and maximum length, and develop a constrained trajectory optimization method to co-optimize the elastic parameters and motion trajectories toward enhanced jumping distance. Results reveal that a less stiff spring is likely to facilitate jumping performance not as a direct propelling source but as a means to unleash more motor power for propelling by trading-off overall energy efficiency. We also visualize the impact of spring coefficients on the overall optimization routine from energetic perspectives to identify the suitable parameter region.
%
\end{abstract}

\section{Introduction}
Contemporary legged robots steadily improve to achieve some impressive locomotion skills. Several approaches, primarily optimization-based, have been found effective to stabilize diverse gaits~\cite{hutter2016anymal,hutter2017anymal,bledt2018cheetah,di2018dynamic,raibert2008bigdog,gehring2013control,lee2020learning,bellicoso2018dynamic} and achieve fast running for quadrupedal robots~\cite{park2017high,katz2019mini,talebi2001quadruped}. 
Achieving \emph{agile} locomotion is also critical to improve the robot's performance in realistic unstructured and complex environments~\cite{hwangbo2019learning,gehringoptimization}.
Examples of high-agility skills in legged robots include rapid turning~\cite{katz2019mini,palmer2010intelligent,yim2018precision}, acrobatics such as back flips~\cite{katz2019mini,kau2019stanford}, and high jumping~\cite{kau2019stanford,nguyen2019optimized,haldane2016robotic,yim2018precision}. 

Standing long jump is also another important high-agility skill, as it can significantly enhance the capability of a legged robot to overcome wide gaps or obstacles~\cite{wong1995control}. 
Existing quadrupedal robots actuated via electric motors can achieve solid jumping performance by relying mostly on rigid body dynamics and powerful actuators at joints (especially Direct-Drive and Quasi-Direct-Drive motors)~\cite{katz2019mini,kenneally2016design,seok2014design,ding2017design,elery2020design}. 
%
Relative to their scale, these quadrupeds feature leg designs that have been optimized over the recent years to exert large ground reaction forces~\cite{katz2019mini,kau2019stanford,blackman2016gait,haldane2017repetitive,arm2019spacebok}, and employ actuators that are at or very near to the most optimal possible selection in terms of output torque density~\cite{katz2019mini,seok2014design,kau2019stanford,hutter2017anymal}. 
Hence, further improving the jumping performance of quadrupeds (e.g., in terms of distance covered in a single long jump) may require rethinking of how the main body of the robot is designed. 

Quadrupedal robot body design that is optimized for long jumping may draw from biological inspiration, especially from jumping masters like Bobcats and Lynxes. 
These animals harness morphological changes over their torso to propel themselves forward during each jump~\cite{hansen2007bobcat}. They can jump over distances several times their body length with elegant manner~\cite{biggins2006bobcat}; this suggests the significance of utilizing both strong leg actuation and \emph{embodied compliance}~\cite{hu2017influence,park2011identification}. 

\begin{figure}[!t]
\vspace{6pt}
\centering
\includegraphics[trim={0cm 0cm 0cm 0cm},clip,width=\linewidth]{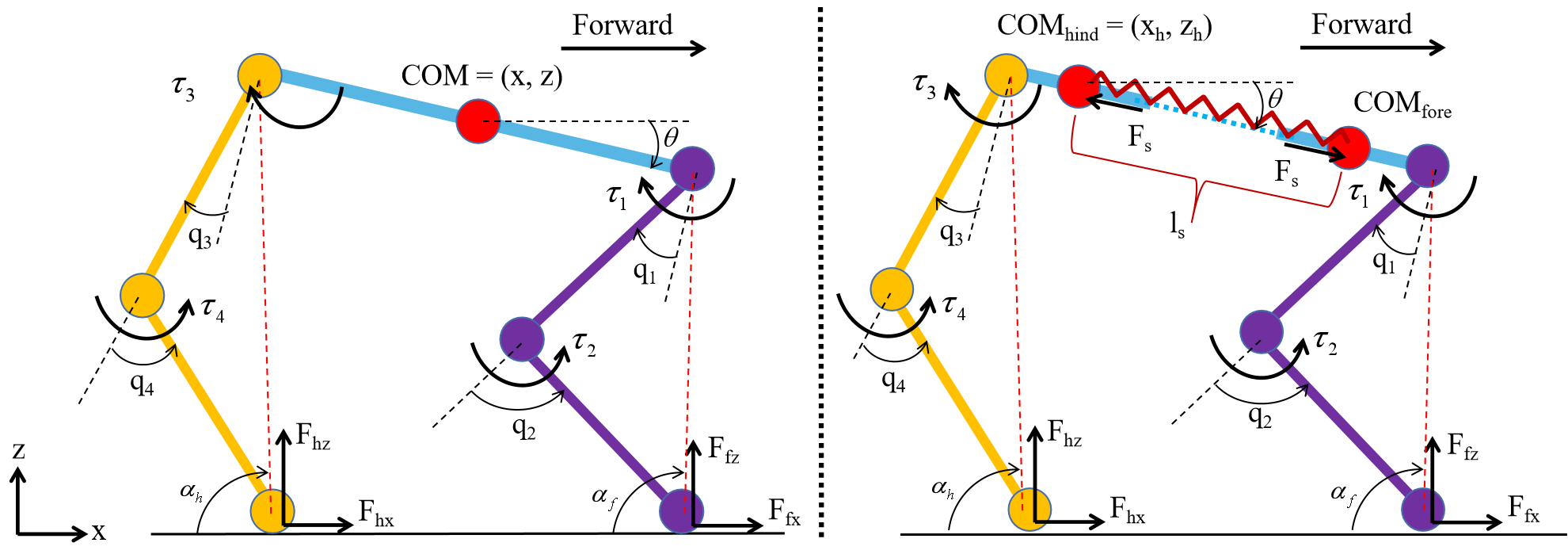}
\vspace{0pt}
\caption{\textbf{Sagittal Plane Models and Posture Descriptor.}
(Left) The reference 5-link rigid model. (Right) Our proposed hybrid model with lock-enabled elastic prismatic spine. 
General coordinates and control inputs are denoted.}
\label{fig:models}
\vspace{-16pt}
\end{figure}

Previous works have demonstrated benefits of embedding spinal compliance for locomotion control. 
For example, the quadrupedal robot Canid~\cite{duperret2016towards} adopts a parallel actuated elastic spine to achieve enhanced jumping compared to rigid body.  Inu~\cite{duperret2017empirical} with the same spine mechanism also achieves stable bounding. 
Lynx-Robot~\cite{eckert2015comparing} uses a compliant segmented spine to self-stabilize gaits like bounding at moderate speeds.  Kitty~\cite{zhao2013spine} employs its flexible spine as a controller to generate different gaits while handling perturbations from varying payloads carried by the robot. 
However, it remains unclear how to best employ compliance and embed it within quadrupedal robot body design optimized for long jumping. 

The motivating question in this paper is how to enable a quadrupedal robot with given leg design and actuators to extend its standing jump length by utilizing a compliant spine. We seek to identify an appropriate spinal compliance setup and develop a trajectory optimization strategy to extend the maximum standing jump distance of the robot. 
To this end, we propose a hybrid model 
with an elastic prismatic spine (Fig.~\ref{fig:models}) that is preloaded by position-controlled servomotor and mechanically locked at initial and maximum length, 
and address 
the co-optimization problem~\cite{ha2018computational,megaro2017computational} of design parameters (i.e. spring constant and rest length) and standing jumping trajectories to improve maximum jumping distance.


To solve these problems, we develop an  
offline nonlinear optimization-based framework
for quadrupedal robots that mainly contributes to 
1) a constrained force-explicit elastic model as a uniform description of the hybrid model, 
2) automatic characterization of suitable elastic parameters, and 
3) trajectory optimization of standing jump with the aim to enhance maximum jumping distance.
Our work also offers insights on the impact of spring coefficients on the overall optimization routine from energetic perspectives, which can be useful to identify the suitable region in parameter space.



\section{Quadruped Dynamic Modeling}\label{sec:Templates}

We focus on sagittal plane dynamics. A 5-link model~\cite{furusho1995realization,nguyen2019optimized} serves as the reference rigid model to compare our proposed hybrid model (Fig.~\ref{fig:models}). 
In the hybrid model, the rigid link-shaped spine of the 5-link model is replaced with a spring-loaded prismatic joint that is preloaded by a servomotor and mechanically locked at predefined initial and final lengths. 
For fair comparison with the rigid model, we set the final locked spine length equal to the spine length of the rigid model; 
the initial length is tunable. Due to the spine locking feature, the hybrid model consists of rigid and elastic sub-models that activate based on a temporal scheduling (see Section~\ref{sec:nonlinearMPC}).  

\subsection{Rigid Sub-model Dynamics}\label{subsec:rigid_full_simp_dynamics}
Articulated quadrupeds with rigid body are typically modeled as floating-base systems~\cite{bellicoso2018dynamic,hutter2012starleth,ma2019first,ma2020bipedal}. We follow this approach here too. 
With reference to Fig.~\ref{fig:models}, the configuration space is 
$\mathbf{\bar{Q}}_r:=\left[x,z,\theta,q_1,q_2,q_3,q_4 \right]^T$, 
with $x$, $z$, and $\theta$ being position and orientation states of spine center of mass (CoM), and 
$\mathbf{q}:= \left[q_1, q_2, q_3, q_4 \right]^T$ 
being the joint angles of fore and hind legs. Ground reaction forces (GRFs) 
and actuator torques at joints are denoted by $\mathbf{F}_{c}:= \left[F_{fx}, F_{fz}, F_{hx}, F_{hz} \right]^T$ and $\mathbf{\tau}:= \left[ \tau_1, \tau_2, \tau_3, \tau_4 \right]^T$, respectively. 
Rigid body dynamics can be derived from Euler-Lagrange equations as 
\begin{align}
    &\mathbf{M}_r(\mathbf{\bar{Q}}_r) \mathbf{\ddot{\bar{Q}}}_r+\mathbf{b}_r (\mathbf{\bar{Q}}_r, \mathbf{\dot{\bar{Q}}}_r)+\mathbf{g}_r (\mathbf{\bar{Q}}_r)=\mathbf{S}^{T} \mathbf{\tau}+\mathbf{J}_{c}^{T} \mathbf{F}_{c} \label{equ:rigid_fullEoM},\\ 
    &\mathbf{J}_{c}(\mathbf{\bar{Q}}_r) \mathbf{\ddot{\bar{Q}}}_r+\dot{\mathbf{J}}_{c}(\mathbf{\bar{Q}}_r) \mathbf{\dot{\bar{Q}}}_r=\mathbf{0}\enspace, \label{equ:kinematic_constraint}
\end{align}
where $\mathbf{M}_r$, $\mathbf{b}_r$ and $\mathbf{g}_r$ are the mass matrix, Coriolis-centrifugal vector and gravitational vector at generalized coordinates, respectively. $\mathbf{J}_{c}$ is the Jacobian of foot locations; selection matrix $\mathbf{S}$ maps $\mathbf{\tau}$ to generalized coordinates $\mathbf{\bar{Q}}_r$.

Rigid-body equations of motion (EoMs)~\eqref{equ:rigid_fullEoM}--\eqref{equ:kinematic_constraint} contain all the important elements of the system. However, high nonlinearity in dynamics and underactuation make it challenging to find an optimal (or even a feasible) solution for the optimization problem, let alone the fact that more constraints are applied. 

To enlarge the reachable 
configuration space, we simplify dynamics~\eqref{equ:rigid_fullEoM}--\eqref{equ:kinematic_constraint} by considering the fact that leg mass is much smaller than that of the main body as shown in Table~\ref{table:model_params}
(this simplification is often used in practice~\cite{raibert1986legged,bledt2018cheetah}), and by treating joint angles $\mathbf{q}$ and 
ground reaction forces $\mathbf{F}_{c}$ as inputs~\cite{bledt2018cheetah}.
Thus, EoMs for the rigid body sub-model become 
\begin{align}
    &\mathbf{M}_r \ddot{\mathbf{Q}}_r+\mathbf{b}_r+\mathbf{g}_r = \mathbf{F}_r(\mathbf{U}_r)\enspace, \label{equ:rigid_EoM}
\end{align}
where $\mathbf{Q}_r:=\left[x,z,\theta\right]^T$, 
the input space is  $\mathbf{U}_r:=\left[ \mathbf{q},\mathbf{F}_{c} \right]^T$, the generalized force is $\mathbf{F}_r = \mathbf{J}_{c,\mathbf{Q}_r}^T \mathbf{F}_c$, and $\mathbf{J}_{c,\mathbf{Q}_r}$ is the Jacobian of contact feet in configuration space.  Further, $\mathbf{M}_r$, $\mathbf{b}_r$ and $\mathbf{g}_r$ are time-invariant, and are given by 
{
\begin{align}
\mathbf{M}_r\hspace{-1pt}&=\hspace{-1pt} \left[ 
\begin{array}{ccc}
m_B & 0 & 0 \\
0 & m_B & 0 \\
0 & 0 & I_B
\end{array}
\right]\hspace{-1pt},\hspace{1pt}
\mathbf{b}_r\hspace{-1pt}=\hspace{-1pt}\mathbf{0},\hspace{1pt}
\mathbf{g}_r\hspace{-1pt}=\hspace{-1pt}
\left[
\begin{array}{c}
0\\
g\,m_B\\
0
\end{array}
\right]\hspace{-1pt}. \label{equ:rigid_EoM_terms}
\end{align}
}%

\begin{table}[!th]
\vspace{6pt}
    \caption{Model Parameters\protect\footnotemark }
    \vspace{-9pt}
    \label{table:model_params}
    \begin{center}
    \renewcommand{\arraystretch}{1.5}
    \begin{tabular}{p{2.4cm}>{\centering}p{1.5cm}>{\centering}p{1.5cm}>{\centering}p{1.5cm}}
        \toprule
        \textbf{Parameter} & \textbf{Symbol} & \textbf{Value} & \textbf{Units} \\
        \midrule
        \midrule
        Body Mass
        & $m_B$ & 22.5 & kg \\ 
        & $m_f$, $m_h$ & 11.25 & kg \\ 
        \midrule
        Body Inertia
        & $I_B$ & 1.05 & $\text{kg} \cdot \text{m}^2$ \\ 
        & $I_f$, $I_h$ & 0.06 & $\text{kg} \cdot \text{m}^2$ \\ 
        \midrule
        Body Length
        & $l_B$ & $0.8$ & m \\ 
        & $l_f$, $l_h$ & $0.2$ & m \\ 
        \midrule
        Leg Mass
        & $m_1$, $m_3$ & 0.4 & kg \\ 
        & $m_2$, $m_4$ & 0.28 & kg \\ 
        Leg Inertia
        & $I_1$, $I_3$ & 0.004 & $\text{kg} \cdot \text{m}^2$ \\ 
        & $I_2$, $I_4$ & 0.0028 & $\text{kg} \cdot \text{m}^2$ \\ 
        Leg length
        & $l_1$, $l_2$, $l_3$, $l_4$ & 0.34 & m \\ 
        \midrule
        Spring Length
        & $l_{s,max}$ & 0.6 & m \\ 
        & $l_{s,min}$ & 0.4 & m \\ 
        \midrule
        Max Joint Torque
        & $\tau_{max}$ & 230 & $\text{N} \cdot \text{m}$ \\ 
        Max Joint Velocity
        & $\dot{q}_{max}$ & 21 & rad/s \\ 
        \bottomrule
    \end{tabular}
    \end{center}
    \vspace{-12pt}
\end{table}

\footnotetext{The parameters are selected based on MIT Cheetah 3~\cite{bledt2018cheetah} for a more realistic physical background. The peak joint torque is set to $0.8 \tau_{max}$ for actuator safety.} 

\subsection{Elastic Sub-model Dynamics}\label{subsec:elastic_full_simp_dynamics}
The major difference between the rigid and elastic sub-models is the addition of the elastic prismatic joint. Yet, 
the system can still be modeled using floating-base coordinates 
if either half of the body is regarded as the floating base~\cite{fisher2017effect}. 

For the elastic sub-model, the setup of states and inputs is similar to the rigid sub-model,  
except that the configuration space is expanded to 
$\mathbf{\bar{Q}}_s:=\left[x_h,z_h,\theta,l_s,q_1,q_2,q_3,q_4 \right]^T$, 
with $x_h$, $z_h$, $\theta$ the position and orientation of hind torso CoM, and $l_s$ the spring length. The full Euler-Lagrange EoMs are in form~\eqref{equ:rigid_fullEoM}--\eqref{equ:kinematic_constraint} 
but with all terms modified to account for new generalized coordinates $\mathbf{\bar{Q}}_s$. 
%
%
%
We can derive a reduced-order dynamical model for the compliant body using the same considerations used to derive~\eqref{equ:rigid_EoM}; however, in this case the compliant force can be described either implicitly or explicitly. In the implicit approach, spring dynamics is included through a spring potential energy term during the Euler-Lagrange formulation. The explicit method regards spring force $F_s$ as input constrained by Hooke's Law. 


Reduced-order dynamics for the elastic body sub-model employs configuration $\mathbf{Q}_s:=\left[ x_h,z_h,\theta, l_s \right]^T$ and is given by 
\begin{align}
    &\mathbf{M}_s(\mathbf{Q}_s) \ddot{\mathbf{Q}}_s+\mathbf{b}_s(\mathbf{Q}_s, \dot{\mathbf{Q}}_s)+\mathbf{g}_s(\mathbf{Q}_s) = \mathbf{F}_s(\mathbf{U}_s)\enspace. \label{equ:springy_EoM_implicit}
\end{align}
In the force-implicit elastic model, $\mathbf{U}_s:=\left[ \mathbf{q},\mathbf{F}_{c} \right]^T$, $\mathbf{F}_s = \mathbf{J}_{c,\mathbf{Q}_s}^T \mathbf{F}_c$, $\mathbf{J}_{c,\mathbf{Q}_s}$ is the Jacobian of contact feet, and $\mathbf{M}_s$, $\mathbf{b}_s$ and $\mathbf{g}_s$ are time-varying and given by
\vspace{-1pt}
{
\begin{align}
\mathbf{M}_s\hspace{-1pt}&=\hspace{-1pt}\left[
\begin{array}{cccc}
m_b & 0 & -m_f\,s_{\theta}\,l_s  & m_f\,c_{\theta}\\
0 & m_b & -m_f\,c_{\theta}\,l_s  & -m_f\,s_{\theta}\\
-m_f\,s_{\theta}\,l_s  & -m_f\,c_{\theta}\,l_s  & m_f\,{l_s  }^2 +I_b & 0\\
m_f\,c_{\theta} & -m_f\,s_{\theta} & 0 & m_f
\end{array}
\right]\hspace{-1pt}, \nonumber \\
\mathbf{b}_s\hspace{-1pt}&=\hspace{-1pt}
\left[
\begin{array}{c}
-m_{f}\,\dot{\theta}  \,{\left(2\, s_{\theta}\,{\dot{l} }_s  + c_{\theta}\,l_s  \,\dot{\theta}  \right)}\\
m_{f}\, s_{\theta}\,l_s  \,{\dot{\theta}  }^2 -2\,m_{f}\, c_{\theta}\,{\dot{l} }_s  \,\dot{\theta}  \\
2\,m_{f}\,{\dot{l} }_s  \,l_s  \,\dot{\theta}  \\
-m_{f}\,l_s  \,{\dot{\theta}  }^2 -k_{s}\,l_{s,\textit{rest}} +k_{s}\,l_s  
\end{array}
\right], \nonumber \\
\mathbf{g}_s\hspace{-1pt}&=\hspace{-1pt}
\left[\quad
0, \quad
g m_b, \quad
-g m_f c_{\theta} l_s, \quad 
-g m_f s_{\theta}
\quad \right]^{T}\enspace.
\label{equ:springy_EoM_implicit_terms}
\end{align}
}%
The total torso mass is $m_b = m_f+m_h$, the sum of inertia is $I_b = I_{f}+I_{h}$, and $sin(\theta)$ and $cos(\theta)$ are abbreviated as $s_{\theta}$ and $c_{\theta}$; all other parameters are listed in Table~\ref{table:model_params}.

In the force-explicit elastic model, 
$\mathbf{U}_s:=\left[ \mathbf{q},\mathbf{F}_{c},F_s \right]^T$ with scalar spring force $F_s$ determined by $F_s = k_s (l_{s,\textit{rest}} - l_s)$, $\mathbf{F}_s = \mathbf{J}_{c,\mathbf{Q}_s}^T \mathbf{F}_c + \left[ 0, 0, 0, F_s \right]^T$, $\mathbf{M}_s$ and $\mathbf{g}_s$ are same as those~\eqref{equ:springy_EoM_implicit_terms}, and 
$\mathbf{b}_s$ is adjusted to 
{
\begin{equation}
    \mathbf{b}_s = 
\left[
\begin{array}{c}
-m_{f}\,\dot{\theta}  \,{\left(2\, s_{\theta}\,{\dot{l} }_s  + c_{\theta}\,l_s  \,\dot{\theta}  \right)}\\
m_{f}\, s_{\theta}\,l_s  \,{\dot{\theta}  }^2 -2\,m_{f}\, c_{\theta}\,{\dot{l} }_s  \,\dot{\theta}  \\
2\,m_{f}\,{\dot{l} }_s  \,l_s  \,\dot{\theta}  \\
-m_{f}\,l_s  \,{\dot{\theta}  }^2
\end{array}
\right]\enspace.
\label{equ:springy_EoM_explicit_terms}
\end{equation}
}%


\textbf{Remark 1. } 
To help distinguish between the implicit and explicit methods, we note that there can be a collision that results in the instantaneous velocity shift right before or after the spring's mechanical lockup, because the two half bodies may have relative velocity along the spine when lockup happens. 
In the implicit way, there is a discrete switching between rigid and elastic sub-models when the spring locks up, along with
the velocity shift that can be computed by conservation of momentum. 
The explicit method, however, is more uniform to describe such hybrid EoMs. During rigid mode, the elastic model can be utilized to represent the rigid model with additional constraints by setting the spring length changing rate $\dot{l}_s$ to zero. During elastic mode, Hooke's Law is enforced onto the spring force. It is observed from our implementation that even though the explicit method has more constraints applied than the implicit way, its uniform description of EoMs makes it more efficient to solve the trajectory optimization problem as we explain next. Hence, in the following we adopt the force-explicit elastic model.


\section{Nonlinear Trajectory Optimization Architecture}\label{sec:nonlinearMPC}
We aim to co-optimize the spring parameters and motion trajectories of the hybrid system represented by the force-explicit elastic model to maximize the standing jump distance, with control horizon targeted at takeoff.

\subsection{Control Horizon}\label{subsec:control_horizon}
\begin{figure}[!t]
	\vspace{6pt}
	\centering
\includegraphics[trim={0cm 0cm 0cm 0cm},clip,width=\linewidth]{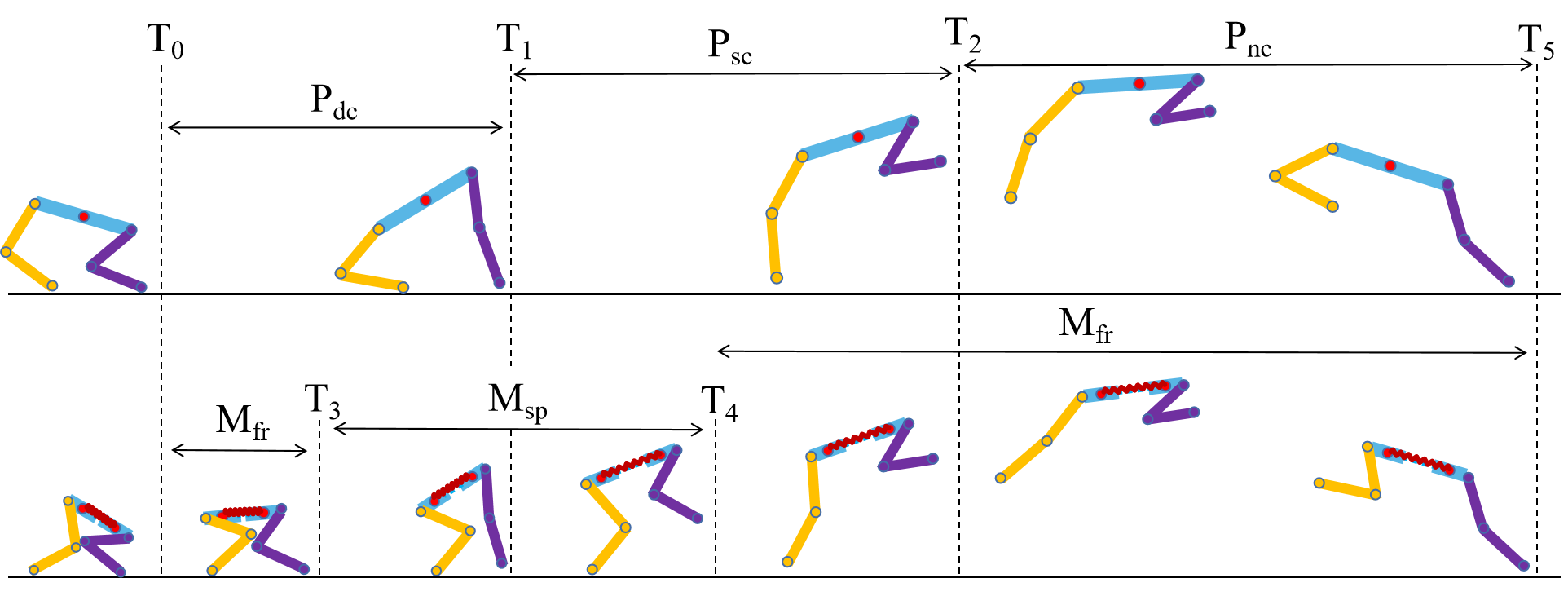}
	\vspace{0pt}
	\caption{\textbf{Phases of Standing Jump Process and their Temporal Scheduling.} A standing jump comprises double-contact $P_{dc}$, single-contact $P_{sc}$, and no-contact $P_{nc}$ phases before landing at $T_5$ for both models, switched by events as fore-foot-liftoff at $T_1$ and hind-foot-liftoff at $T_2$. Takeoff happens during $T_0$ to $T_2$. The proposed model (shown in bottom row) also experiences rigid mode $M_{fr}$ and elastic mode $M_{sp}$, switched by events as spring-release at $T_3$ and spring-lock at $T_4$. Hind legs can be arranged as knee-backward (top) or knee-forward (bottom), which is discussed in Section~\ref{subsect:prelim_results}.}
	\label{fig:jump_phases_modes}
	\vspace{-16pt}
\end{figure}

As shown in Fig. \ref{fig:jump_phases_modes}, a standing jump consists of three phases, namely takeoff, flight, and landing. 
By ignoring the leg dynamics, the parabolic trajectory $x(t)$ and $z(t)$ of body CoM during flight phase is determined by translational velocities $\dot{x}(T_2)$ and $\dot{z}(T_2)$ at the end of takeoff phase, or $\dot{x}(T_5)$ and $\dot{z}(T_5)$ at the beginning of landing phase.

The symmetric principle can be applied to enforce trajectories of rotational states $\theta (t)$ and $\dot{\theta} (t)$ such that the body pitch angle $\theta (t)$ is opposite between instants $T_2$ and $T_5$. The symmetric principle is also used in the literature to consider takeoff and landing reflected in every way~\cite{wong1995control}. 
The purpose of this present paper is to identify possible jumping distance upper-bound(s) determined by either $\dot{x}(T_2)$ and $\dot{z}(T_2)$, or $\dot{x}(T_5)$ and $\dot{z}(T_5)$, all four of which, however, are not known. For landing phase, $\dot{x}(T_5)$ and $\dot{z}(T_5)$ are initial conditions and their being unknown makes the nonlinear trajectory optimization problem very hard to tackle. On the other hand, $\dot{x}(T_2)$ and $\dot{z}(T_2)$ are final terms in takeoff phase and can be indicated in the cost function in a straightforward manner. Therefore, we choose to optimize trajectories for takeoff phase.

During takeoff phase,
Fig. \ref{fig:jump_phases_modes} also illustrates several important events that switch the model dynamics within finite types: contact switching events like fore foot lifting off and hind foot lifting off, and mode switching events like spine spring releasing and locking.
Note that the effect of various temporal scheduling lies beyond the scope of this paper and is part of future work. 

\subsection{Rigid Model Standing Jump Evaluation}\label{subsec:rigid_jump_eval}
We first estimate the maximum standing jump distance $D_r$ with rigid model under specified timeline setup and initial posture conditions.
We preset trigger time of contact switching events in the timeline as $T_1$ and $T_2$ in Fig. \ref{fig:jump_phases_modes}. 
The initial posture is defined by simplifying the 5-link model to a 3-link model (See Fig. \ref{fig:models}): each 2-segment leg is replaced by a 1-segment virtual leg that links hip joint and foot; the initial posture is defined by torso position and orientation $x^{init}$, $z^{init}$, $\theta^{init}$, and virtual leg attacking angles $\alpha_f^{init}$ and $\alpha_h^{init}$; the initial positions for leg joints $\mathbf{q}$ are then derived by inverse kinematics.

The control horizon is set to be $p$ time steps, with each time step of duration $T_s$ so that $T_2 = T_s p$. Then, the continuous-time nonlinear trajectory optimization problem is organized over $\left[ T_0, T_2\right]$ as 
\begin{align}
\begin{array}{l}
\min _{\mathbf{X}, \mathbf{U}} J(\mathbf{X})\\
\text {subject to } \dot{\mathbf{X}} = \mathbf{f}(\mathbf{X},\mathbf{U}) \\
\mathbf{C}_{eq}(\mathbf{X},\mathbf{U}) = \mathbf{0} \\
\mathbf{C}_{ineq}(\mathbf{X},\mathbf{U}) \leq \mathbf{0}
\end{array}\label{equ:rigid_nlmpc}
\end{align}
where states $\mathbf{X}=\left[ \mathbf{Q}_r, \dot{\mathbf{Q}}_r, t \right]^T$,
$t$ is the time, inputs $\mathbf{U} = \mathbf{U}_r$, 
and dynamic constraints $\dot{\mathbf{X}} = \mathbf{f}(\mathbf{X},\mathbf{U})$ are based on \eqref{equ:rigid_EoM} as
\begin{align}
\mathbf{f}
=\left[
\begin{array}{c}
\dot{\mathbf{Q}}_r \\ 
\mathbf{M}_r^{-1}(\mathbf{F}_r(\mathbf{U}_r) - \mathbf{b}_r - \mathbf{g}_r) \\
1
\end{array}
\right]\enspace.
\end{align}

Standard form for cost function needs reference trajectories for states and also minimizes the input effort~\cite{bledt2018cheetah,nguyen2019optimized}. However, preset desired state trajectories may not lead the system reaching the maximum jumping distance. Therefore, we directly reflect the jumping distance $D = 2 \dot{x}(T_2) \dot{z}(T_2) / g$ in cost function as
\begin{align}\label{eq:J}
J = \dot{x}(T_2) \dot{z}(T_2)
\end{align}

Equality and inequality constraints are enforced to satisfy physical limitations of the robot and its interaction with the environment as

\begin{itemize}
    \item Symmetric flight limits: $\dot{\theta}(T_2) = -\frac{g \theta(T_2)}{\dot{z}(T_2)}$
    \item Contact feet limits: stance foot $i$ position $\mathbf{p}_i = \textit{constant}$
    \item Joint angle limits: $\mathbf{q}_{min} \leq \mathbf{q}(t) \leq \mathbf{q}_{max}$
    \item Joint velocity limits: $-\dot{\mathbf{q}}_{max} \leq \dot{\mathbf{q}}(t) \leq \dot{\mathbf{q}}_{max}$
    \item Joint torque limits: $-\mathbf{\tau}_{max} \leq \mathbf{\tau}(t) \leq \mathbf{\tau}_{max}$, $\tau = \mathbf{J}_{c,\mathbf{q}} \mathbf{F}_c$, where $\mathbf{J}_{c,\mathbf{q}}$ is Jacobian of stance feet in joint space
    \item Minimum ground clearance at joint $q_i$: $p_{q_i,z} > z_{min}$
    \item Takeoff posture limits: $\dot{x}(T_2) > 0$, $\dot{z}(T_2) > 0$, $\theta (T_2) < 0$, and $\dot{\theta} (T_2) > 0$
    \item Coulomb friction limits at stance foot $i$: $-\mu \leq \frac{F_{i,x}}{F_{i,z}} \leq \mu$
    \item Minimum normal GRF at stance foot $i$: $F_{i,z} > F_{min}$
    \item Geometric limits for swing legs
\end{itemize}

The solution of the optimization problem gives the longest jumping distance $D_r^{*}$ for the rigid model. 

\subsection{Compliant Model Jumping Trajectories Optimization}
Spring coefficients $k_s$ and $l_{s,rest}$ are unknown and need to be properly selected. However, to our knowledge, there is no direct design principle available to guide that selection. 
Nevertheless, we can treat the spring coefficients $k_s$ and $l_{s,rest}$ as additional constrained inputs in the trajectory optimization problem, and seek to adjust these parameters and evaluate final jumping distance simultaneously.



The pre-configuration for contact switching and initial posture is same as that from Section \ref{subsec:rigid_jump_eval}. Note that the initial posture is set for the whole body CoM instead of the hind half body only. Additionally, the trigger time of mode switching events is preset as $T_3$ and $T_4$. The spring releasing length is set as $l_{s} (T_3) = l_{s} (T_0) = l_{s,min}$. The spring locking length is set as $l_{s} (T_4) = l_{s} (T_2) =  l_{s,max} = l_{B} - 0.5 l_{f} - 0.5 l_{h}$ such that the overall spine after $T_4$ has the same length as in the case of the rigid model spine.

We initialize $k_s(T_0)$ and $l_{s,rest} (T_0)$ with $k_s^{init}$ and $l_{s,rest}^{init}$, and use the same horizon setup as in Section \ref{subsec:rigid_jump_eval} and the same general form of constrained trajectory optimization problem~\eqref{equ:rigid_nlmpc}, with terms adjusted as 
\begin{align}
\mathbf{X} &=\left[ \mathbf{Q}_s, \dot{\mathbf{Q}}_s, t \right]^T \\
\label{equ: U_Fa}
\mathbf{U} &= \mathbf{U}_s=\left[ \mathbf{q},\mathbf{F}_{c},F_s,k_s,l_{s,rest} \right]^T \\ 
\mathbf{f} &=\left[
    \begin{array}{c}
    \dot{\mathbf{Q}}_s \\ 
    \mathbf{M}_s^{-1}(\mathbf{F}_s(\mathbf{U}_s) - \mathbf{b}_s - \mathbf{g}_s) \\
    1
    \end{array}
    \right]
\end{align}
where $\mathbf{M}_s$ and $\mathbf{g}_s$ are from (\ref{equ:springy_EoM_implicit_terms}).
The jumping distance is
\begin{align}
D_s = 2 \dot{x}_c(T_2) \dot{z}_c(T_2) / g \label{equ:Ds_compliance_shaping}
\end{align}
where $\dot{x}_c$ and $\dot{z}_c$ are velocities of torso CoM, and because the rigid model is enforced at $T_2$, they can be derived as
\begin{align}
\left[
\begin{array}{l}
\dot{x}_c \\
\dot{z}_c
\end{array}
\right]
=\left[
\begin{array}{c}
{\dot{x} }_h  -\frac{l_s \,s_{\theta}\,\dot{\theta}  }{2}\\
{\dot{z} }_h  -\frac{l_s \,c_{\theta}\,\dot{\theta}  }{2}
\end{array}
\right] \label{equ:pdot_c}
\end{align}
therefore, the cost function is formed as 
\begin{align}
J = \dot{x}_c(T_2) \dot{z}_c(T_2). \label{eq:J2}
\end{align}
In the case that the desired enhanced distance is determined, we can adopt $J = (D_s^{des} - D_s)^2$ as the alternative. In this paper, we focus on $J$ in~\eqref{eq:J2}.

Equality constraints $\mathbf{C}_{eq}(\mathbf{X},\mathbf{U}) = \mathbf{0}$ and inequality constraints $\mathbf{C}_{ineq}(\mathbf{X},\mathbf{U}) \leq \mathbf{0}$ from Section \ref{subsec:rigid_jump_eval} are enforced with additional constraints as
\begin{itemize}
    \item Hooke' Laws on spring force: $F_s(t) = k_s(t) (l_{s,rest}(t)-l_s(t)), t \in [T_3, T_4]$;
    \item Spring coefficients limits: $k_s(t) = k_s(T_3) > 0$, $l_{s,rest}(t) = l_{s,rest}(T_3) \geq l_{s,max}$, $t \in [T_3, T_4]$;
    \item Free spring limits: $ l_{s,min} \leq l_s(t) \leq l_{s,max}, t \in [T_3, T_4]$;
    \item Locked spring limits: $ l_s(t) = l_{s,min}, t \in [T_0, T_3]$, and $ l_s(t) = l_{s,max}, t \in [T_4, T_2]$.
\end{itemize}

The optimization solutions to the trajectories of system states and inputs are described as
\begin{align}
\begin{split}
    \mathbf{Q}_s^{*}:&=\left[ x_h^{*},z_h^{*},\theta^{*}, l_s^{*} \right]^T, \quad
    \dot{\mathbf{Q}}_s^{*}:=\left[ \dot{x}_h^{*},\dot{z}_h^{*},\dot{\theta}^{*}, \dot{l}_s^{*} \right]^T \\
    \mathbf{q}^{*}:&=\left[ q_1^{*},q_2^{*},q_3^{*}, q_4^{*} \right]^T, \quad \dot{\mathbf{q}}^{*}:=\left[ \dot{q}_1^{*},\dot{q}_2^{*},\dot{q}_3^{*}, \dot{q}_4^{*} \right]^T \\
    \mathbf{\tau}^{*}:&=\left[ \tau_1^{*},\tau_2^{*},\tau_3^{*}, \tau_4^{*} \right]^T \label{equ:nominal_traj}
\end{split}
\end{align}
where joint velocities $\dot{\mathbf{q}}^{*}$ are evaluated based on $\mathbf{q}^{*}$ and time step $T_s$. The solution also provides the spring coefficients $k_s^{*}$ and $l_{s,rest}^{*}$ refined from their initial values.

\begin{figure*}[!t]
\vspace{6pt}
\centering
\includegraphics[trim={3.5cm 0.0cm 3.0cm 0cm},clip,width=0.95\linewidth]{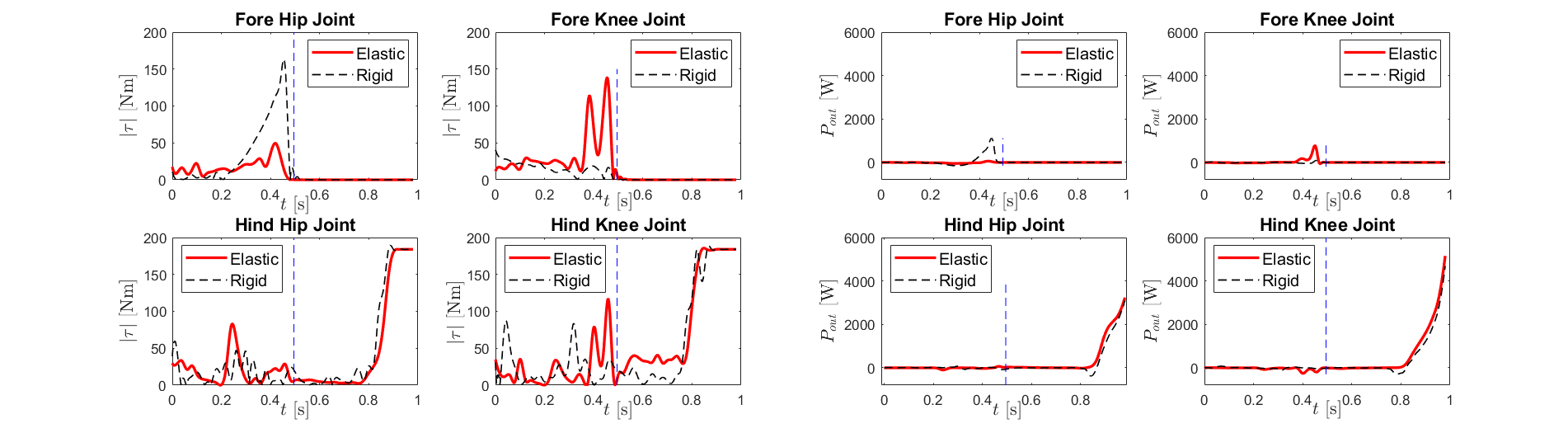}
\vspace{-10pt}
\caption{\textbf{Energy Consumption During Takeoff.} (Left four panels) Trajectories of motor torque amplitude at each joint. (Right four panels) Progress of output power at each joint. The switching time at fore foot liftoff event is indicated as blue line.}
\label{fig:energy_consumption}
\vspace{-15pt}
\end{figure*}

\section{Results and Discussion}\label{sec:Results}
\subsection{Choices of Spring Initial Conditions}\label{subsect:initial_conditions}
The optimizing solver with different initial conditions of the spring parameters is likely to produce distinct results. To better identify the reasonable parameter space, we approach the problem from two perspectives. 
On the aspect of engineering feasibility, the commercially available servomotors to preload the spring confine the spring parameter space. The limited space on the robot body requires the servomotor to be compact. To our knowledge, most compact servomotors (e.g., Robotis' Dynamixel) offer stall torque less than 10 Nm, and with lever length close to 5 cm and spring compressed length near 0.5 m, the safe operational pulling force may be less than 150 N and the spring constant less than 300 N/m. Note that the springs in the above parameter space are not likely to store potential energy more than 50 J, a rather small amount compared to total motor output work of the rigid model that is more than 500 J. 
This observation implies that the above parameter space may contribute to less stiffer springs.
On the other hand, if the spring is assumed to be the major energy source for the jumping improvement over $20\%$, the spring needs to store more than 100 J energy, accounting for the spring constant close to 1000 N/m under the similar estimation logic as above. In fact, the utilization of the spring potential energy could be less efficient and it requires much stiffer springs, altogether accounting for more servo effort and weight.
To alleviate the engineering challenge, we focus on less stiff springs in the tests and leave stiffer spring setups as part of future research.

\begin{figure}[!t]
	\vspace{4pt}
	\centering
\includegraphics[trim={1cm 0.55cm 1.0cm 0.9cm},clip,width=0.85\linewidth]{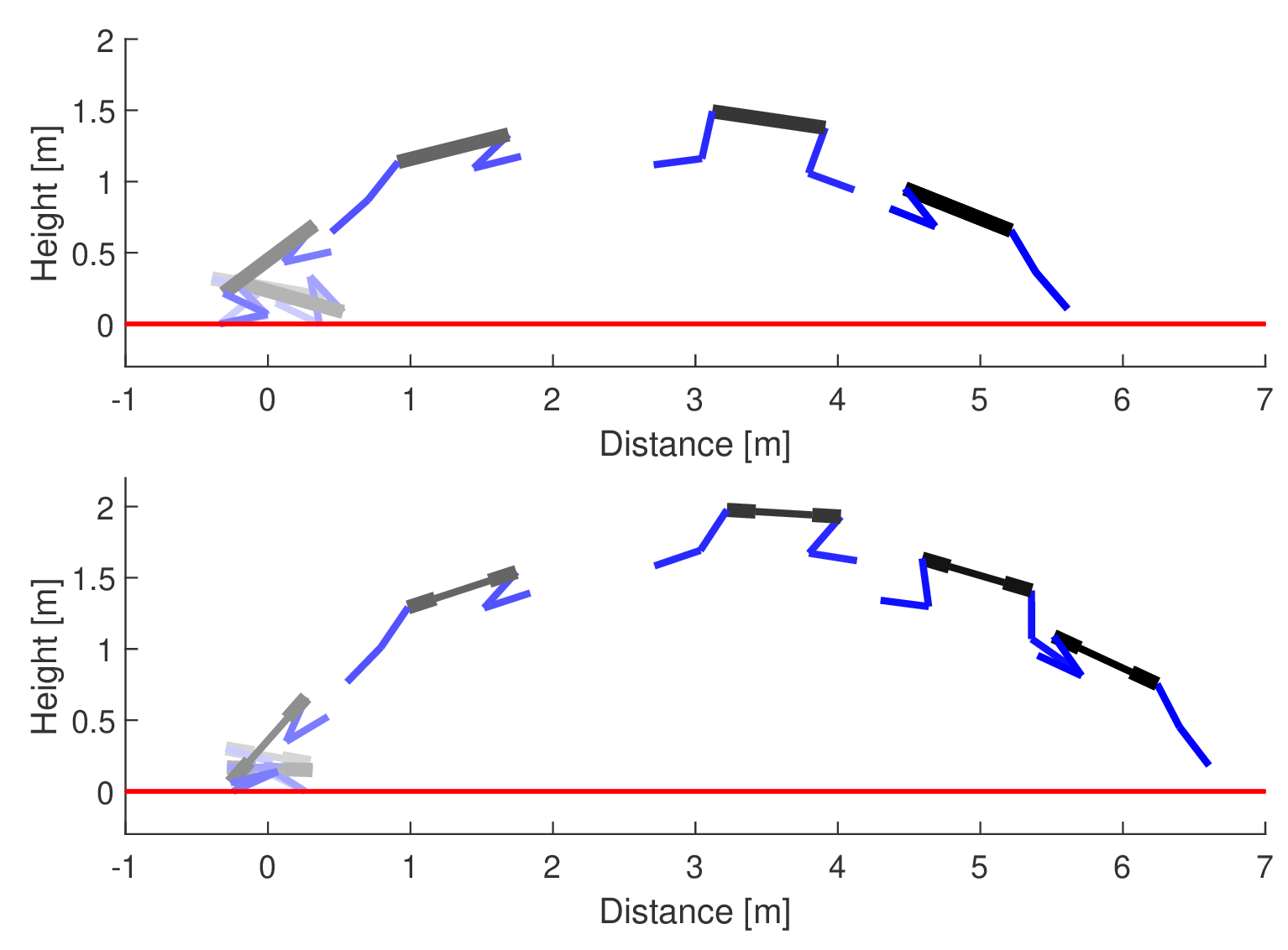}
	\vspace{-1pt}
	\caption{\textbf{Standing Jump Sequence.} Jumping trajectories during takeoff and flight for rigid (top) and hybrid (bottom) models with a sample optimized setup ($k_s^{*} = 180 \text{ N/m}$ and $l_{s,rest}^{*} = 0.7832 \text{ m}$). Ground is shown as red line. (Figure best viewed in color.)}
	\label{fig:jump_sequence}
	\vspace{-16pt}
\end{figure}

\subsection{Standing Long Jumping Results}\label{subsect:prelim_results}
We tested distinct setups for phase temporal scheduling and initial posture; the majority of tests achieved over $10\%$ and many achieved over $20\%$ extension of jumping distance. 
Figure~\ref{fig:jump_sequence} depicts rigid and hybrid model jumping for a sample optimized setup. 
For both models 
$\left[ T_0, T_1, T_2, T_3, T_4 \right] = \left[ 0, 0.5, 1.0, 0.4, 0.8 \right]\text{ s}$ and 
$[ x^{init},z^{init},\theta^{init},\alpha_f^{init},\alpha_h^{init} ] = \left[ 0 \text{ m}, 0.25 \text{ m}, 10^{\circ}, 100^{\circ}, 80^{\circ} \right]$,
with $T_s=0.02$\;s. 
The rigid model jumps for $4.5545$\;m and the hybrid model exceeds by $23\%$ and reaches $5.6086$\;m. The spring parameters are optimized as $k_s^{*} = 180$\;N/m and $l_{s,rest}^{*} = 0.7832$\;m,
with initial guess as $k_s^{init} = 153 \text{ N/m}$ and $l_{s,rest}^{init} = 0.7499 \text{ m}$.

We compared energy consumption between models. 
Motor output power can be evaluated as $P_{out} = \tau \cdot \omega$, with $\tau$ as motor shaft torque and $\omega$ as shaft angular velocity. As shown in Fig. \ref{fig:energy_consumption}, both models exhibit a similar motor pattern. The motors at fore leg joints lead major work on posture adjustment before fore foot lifting up, followed by the motors at hind leg joints offering major contribution to propelling the robot up-forward. In the double contact phase, both models have similar peak torque on the fore leg but for different joints. 
In the single contact phase, 1) both models experience a smooth output power increase on the hind leg within the last $0.2$\;s rather than an abrupt step-up; 
2) the hybrid model spring is locked during the propelling phase, implying that the spring contributes to the preparation phase before propelling;
3) the rigid model experiences some negative motor work right before propelling at about $0.8$\;s, while the hybrid model has a smoother startup; 
4) based on these, 
the hybrid model exerts more motor power at both hip and knee joints on the hind leg during propelling.
The above observations suggest that less stiff springs do not contribute to propelling the body directly but instead assist the system reach a better startup status right before propelling.


\begin{figure*}[!t]
	\vspace{6pt}
	\centering
    \includegraphics[trim={1cm 0cm 1cm 0cm},clip,width=0.49\linewidth]{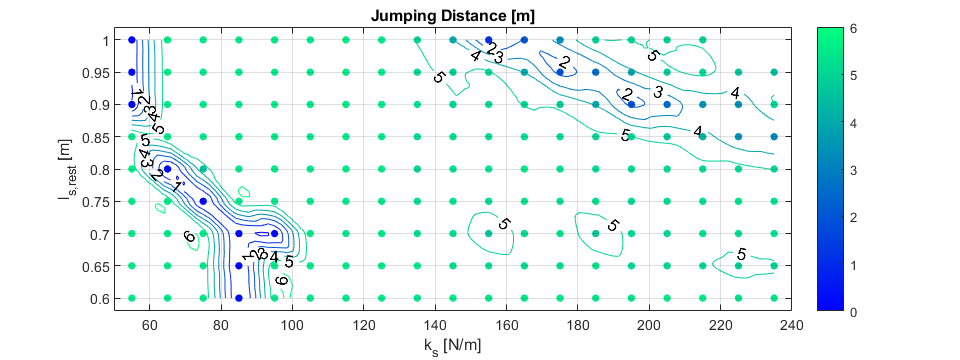}
	\vspace{0pt}
	\includegraphics[trim={1cm 0cm 1cm 0cm},clip,width=0.49\linewidth]{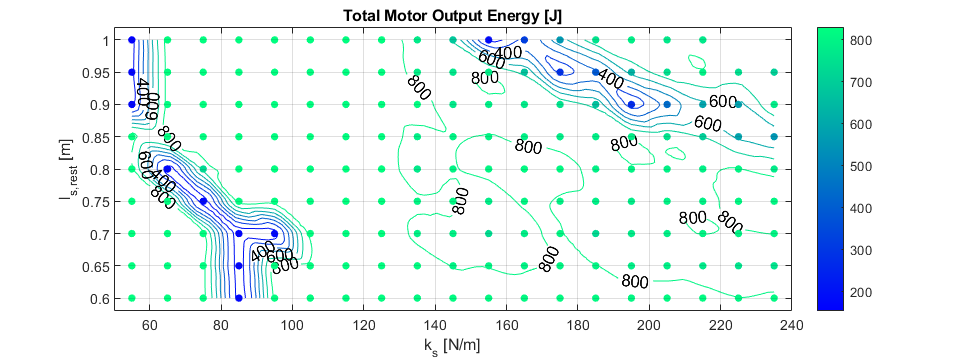}
	\vspace{0pt} \\
	\includegraphics[trim={1cm 0cm 1cm 0cm},clip,width=0.49\linewidth]{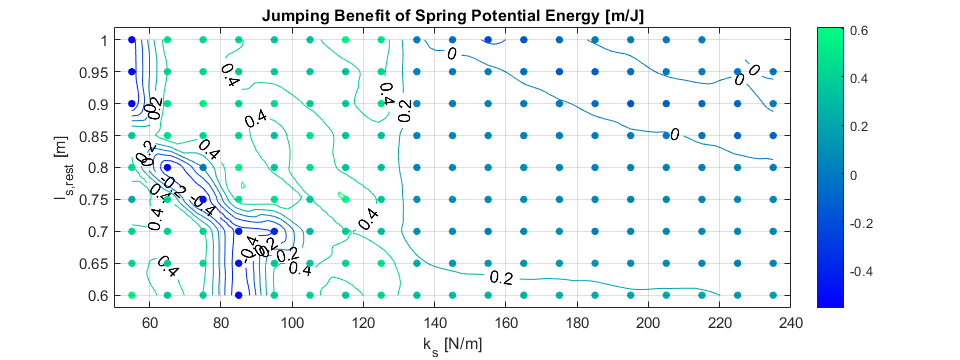}
	\vspace{0pt}
	\includegraphics[trim={1cm 0cm 1cm 0cm},clip,width=0.49\linewidth]{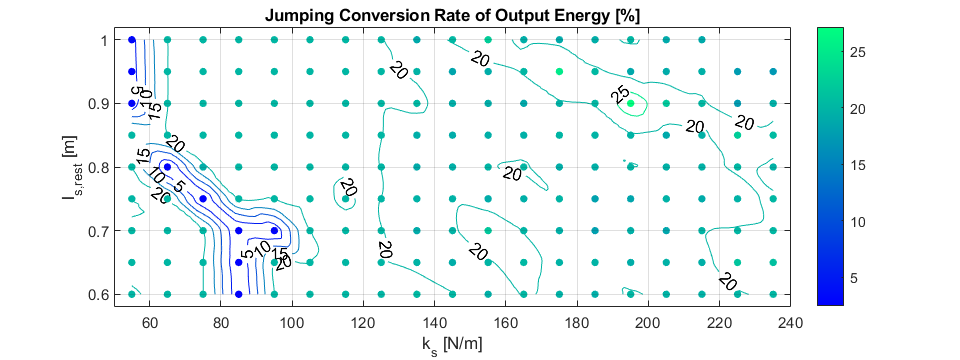}
	\vspace{-3pt}
	\caption{\textbf{Impact of Elastic Characteristics.} Relation between spring parameters ($k_s$, $l_{s,rest}$) and indicators of interest. Collected data are down-sampled and presented in the colored scatters. Contours are generated from the original database.}
	\label{fig:characteristics_analysis}
	\vspace{-15pt}
\end{figure*}

As indicated in Fig. \ref{fig:jump_phases_modes}, we adopted two arrangements for hind leg as knee-backward and knee-forward.
The knee-backward arrangement appears more often for aggressive motion in the literature~\cite{katz2019mini,nguyen2019optimized} and in commercial products (like Boston Dynamics's Spot, Ghost Robotics' Vision and Spirit, and Unitree's Laikago and A1), and the knee-forward arrangement offers high-performance locomotion~\cite{grimminger2020open,hutter2017anymal}. 
Results obtained herein, appear to suggest that knee-forward hind legs are more likely to produce better solutions with less computing time even though both arrangements should be dynamically identical in principle. Shedding more light in this direction is part of future work, building upon recent related findings that suggest the energetic efficiency of the elbows-back, knees-forward arrangement~\cite{usherwood2020limb}.

\subsection{Effect of Spring Coefficients}\label{subsect:spring_coefficients}
It was observed that different spring initial conditions did not guide the optimization toward the same optimal solution or, at least, to a tight region of solutions. Some solutions have similar jumping distance but distinct results of the spring coefficients, implying the nonconvexity of the optimization problem. This finding suggests that optimizing for elastic parameters $k_s$ and $l_{s,rest}$ using some random baseline may lead to suboptimality. 
To visualize the limited region of the coefficients that can benefit the system, we used the setup from Section~\ref{subsect:prelim_results} and searched the parameter space of less stiff springs ($k_{s} = 50 \sim 240 \text{ N/m}$ and $l_{s,rest} = 0.6 \sim 1.0 \text{ m}$) explained in Section \ref{subsect:initial_conditions}.
Each sampled parameter pair characterized a specific spring and was fixed into the model to solve for a solution.

Results are presented in Fig. \ref{fig:characteristics_analysis}.
The upper left graph shows that the springs of moderate level in the less stiff range are more likely to produce jumping distance larger than $5$\;m (that is, at least $10\%$ improvement). Increasing or decreasing both stiffness and rest length may risk undermining the performance.
In the upper right graph, we demonstrate the total motor output energy $E_m$ calculated by the sum of their mechanical work. The similarity of the boundary pattern is noticeable between contours in the upper left and right graphs. This indicates that motor power enforces direct effort onto the jumping distance and it is still the major source in the hybrid model for the improvement of jumping performance. On the other hand, it also reflects that a suitable less stiff spring can help unleash more power from the motors, as specifically explained in Section \ref{subsect:prelim_results}.
In the lower left graph, we illustrate the parameter effect on the jumping benefit $B$ of the spring potential energy, defined by the jumping distance improvement $\delta D$ over preloaded spring potential energy $E_{s}$ (i.e. $B = \delta D / E_{s}$). This indicator also keeps the boundaries from the upper two graphs but segments the area of the moderate level into several sub-areas, some of which suggest better jumping performance with less spring energy preloaded.
The lower right graph aims to exhibit the situation of energy efficiency reflected by the conversion rate $CVR$ from total output energy $E$ to the kinematic energy $E_{k}$ at $T_2$ (i.e. $CVR = E_{k}(T_2) / E$, $E = E_m + E_s$). Conventionally, the total input power is required to compute $CVR$, but it is not available since there is no specific assumption of motor power curve. Therefore, we adopt motor output power for the calculation, which will lightly increase the $CVR$ though we expect the contour pattern to be similar. From the results, we observe an energy efficiency drop with the use of less stiffer springs. (For reference, the rigid model efficiency is $31.02\%$.)

\section{Conclusions}
The paper contributes an optimization-based framework for modeling and trajectory planning of quadrupedal robot standing jump. A preloaded lock-enabled elastic model is introduced for the spine, and a constrained trajectory optimization method is proposed for co-optimization of spring parameters and motion trajectories. Results suggest that less stiff springs can help unleash more motor power and eventually improve jumping performance.
Future directions of research include 1) study of the effect of much stiffer springs, and 2) realization of the preloaded elastic spine on a physical quadrupedal robot.


	\bibliographystyle{IEEEtran}
	\bibliography{reference.bib}
	
	

\end{document}